\documentclass[sigconf]{acmart}
\AtBeginDocument{%
  \providecommand\BibTeX{{%
    \normalfont B\kern-0.5em{\scshape i\kern-0.25em b}\kern-0.8em\TeX}}}

\copyrightyear{2023} 
\acmYear{2023} 
\setcopyright{acmlicensed}\acmConference[KDD '23]{Proceedings of the 29th
ACM SIGKDD Conference on Knowledge Discovery and Data Mining}{August
6--10, 2023}{Long Beach, CA, USA}
\acmBooktitle{Proceedings of the 29th ACM SIGKDD Conference on Knowledge
Discovery and Data Mining (KDD '23), August 6--10, 2023, Long Beach, CA,
USA}
\acmPrice{15.00}
\acmDOI{10.1145/nnnnnnn.nnnnnnn}
\acmISBN{xxx-x-xxxx-xxxx-x/YY/MM}

\usepackage[ruled, lined, vlined, linesnumbered]{algorithm2e}
\usepackage{amsmath, amsthm}
\usepackage{bm}
\usepackage{multirow}
\usepackage{nccmath}
\usepackage{subfig}
\usepackage{threeparttable}
\usepackage{xcolor, xspace}
\usepackage{makecell}
\usepackage{algorithmic}
\usepackage{nccmath}

\theoremstyle{definition}

\newcommand{\eg}{\emph{e.g.},\xspace}
\newcommand{\ie}{\emph{i.e.},\xspace}

\newcommand{\wrt}{\emph{w.r.t.}\xspace}
\newcommand\figref[1]{Figure.~\ref{#1}}
\newcommand\tabref[1]{Table~\ref{#1}}
\newcommand\secref[1]{Sec.~\ref{#1}}
\newcommand\equref[1]{Eq.(\ref{#1})}
\newcommand\algoref[1]{Alg.~\ref{#1}}
\newcommand\appref[1]{Appendix~\ref{#1}}
\newcommand{\fakeparagraph}[1]{\vspace{1mm}\noindent\textbf{#1.}}

\newcommand{\sysname}{AGS\xspace}

\ifodd 1

\newcommand{\xx}[1]{{\color{teal}{#1}}}
\newcommand{\TODO}[1]{\textbf{\color{red}{TODO: #1} }}

\else

\newcommand{\xx}[1]{{#1}
\newcommand{\TODO}[1]{#1}
\fi

\begin{document}

\title{Localised Adaptive Spatial-Temporal Graph Neural Network}

\author{Wenying Duan}
\affiliation{%
  \institution{School of Mathematics and Computer Science\\Nanchang University}
  \city{Nanchang}
  \country{China}}
\email{wenyingduan@ncu.edu.cn}

\author{Xiaoxi He}
\authornote{corresponding author}
\affiliation{%
  \institution{Faculty of Science and Technology\\University of Macau}
  \city{Macau}
  \country{China}}
\email{hexiaoxi@um.edu.mo}

\author{Zimu Zhou}
\affiliation{%
  \institution{School of Data Science\\City University of Hong Kong}
  \city{Hong Kong}
  \country{China}}
\email{zimuzhou@cityu.edu.hk}

\author{Lothar Thiele}
\affiliation{%
  \institution{D-ITET\\ETH Zurich}
  \city{Zurich}
  \country{Switzerland}}
\email{thiele@ethz.ch}

\author{Hong Rao}
\affiliation{%
  \institution{School of Software\\Nanchang University}
  \city{Nanchang}
  \country{China}}
\email{raohong@ncu.edu.cn}

\begin{abstract}
Spatial-temporal graph models are prevailing for abstracting and modelling spatial and temporal dependencies. 
In this work, we ask the following question: \textit{whether and to what extent can we localise spatial-temporal graph models?}
We limit our scope to adaptive spatial-temporal graph neural networks (ASTGNNs), the state-of-the-art model architecture.
Our approach to localisation involves sparsifying the spatial graph adjacency matrices. 
To this end, we propose Adaptive Graph Sparsification (\sysname), a graph sparsification algorithm which successfully enables the localisation of ASTGNNs to an extreme extent (fully localisation).
We apply \sysname to two distinct ASTGNN architectures and nine spatial-temporal datasets. Intriguingly, we observe that spatial graphs in ASTGNNs can be sparsified by over 99.5\% without any decline in test accuracy. Furthermore, even when ASTGNNs are fully localised, becoming graph-less and purely temporal, we record no drop in accuracy for the majority of tested datasets, with only minor accuracy deterioration observed in the remaining datasets.
However, when the partially or fully localised ASTGNNs are reinitialised and retrained on the same data, there is a considerable and consistent drop in accuracy.
Based on these observations, we reckon that \textit{(i)} in the tested data, the information provided by the spatial dependencies is primarily included in the information provided by the temporal dependencies and, thus, can be essentially ignored for inference;
and \textit{(ii)} although the spatial dependencies provide redundant information, it is vital for the effective training of ASTGNNs and thus cannot be ignored during training.
Furthermore, the localisation of ASTGNNs holds the potential to reduce the heavy computation overhead required on large-scale spatial-temporal data and further enable the distributed deployment of ASTGNNs.
\end{abstract}
%

\ccsdesc[500]{Computing methodologies~Neural networks}
\keywords{graph sparsification, spatial-temporal graph neural network, spatial-temporal data}

\maketitle

\section{Introduction}
\label{sec:intro}
An increasing number of modern intelligent applications \cite{bib:TNNLS20:Wu, bib:NIPS20:Bai, bib:ICLR22:Chen} rely on \textit{spatial-temporal data}, \ie data collected across both space and time.
Spatial-temporal data often contain \textit{spatial and temporal dependencies}, \ie the current measurement at a particular location has causal dependencies on the historical status at the same and other locations.
Learning these spatial and temporal dependencies is usually the essence of spatial-temporal data mining and plays a vital role in spatial-temporal inference \cite{bib:CSUR18:Atluri}.

Spatial-temporal data can be effectively represented by \textit{spatial-temporal graph models}, which can depict complex relationships and interdependencies between objects in non-Euclidean domains \cite{bib:AAAI18:Yan, bib:AAAI19:Guo, bib:ICONIP18:Seo, bib:ICLR18:Li, bib:IJCAI18:Yu, bib:IJCAI:bai, bib:IJCAI20:Huang, bib:TNNLS20:Wu, bib:TNNLS20:Wu}.
Of our particular interest are \textit{adaptive spatial-temporal graph neural networks} (ASTGNNs), a popular class of spatial-temporal graph models which 
have demonstrated outstanding performance in applications involving spatial-temporal data such as traffic forecasting, blockchain price prediction, and biosurveillance forecasting \cite{bib:IJCAI19:Wu, bib:TNNLS20:Wu, bib:NIPS20:Bai, bib:ICML21:Chen, bib:AAAI22:Choi, bib:ICLR22:Chen}.

\begin{figure}[t]
    \centering
    \includegraphics[width=1\linewidth]{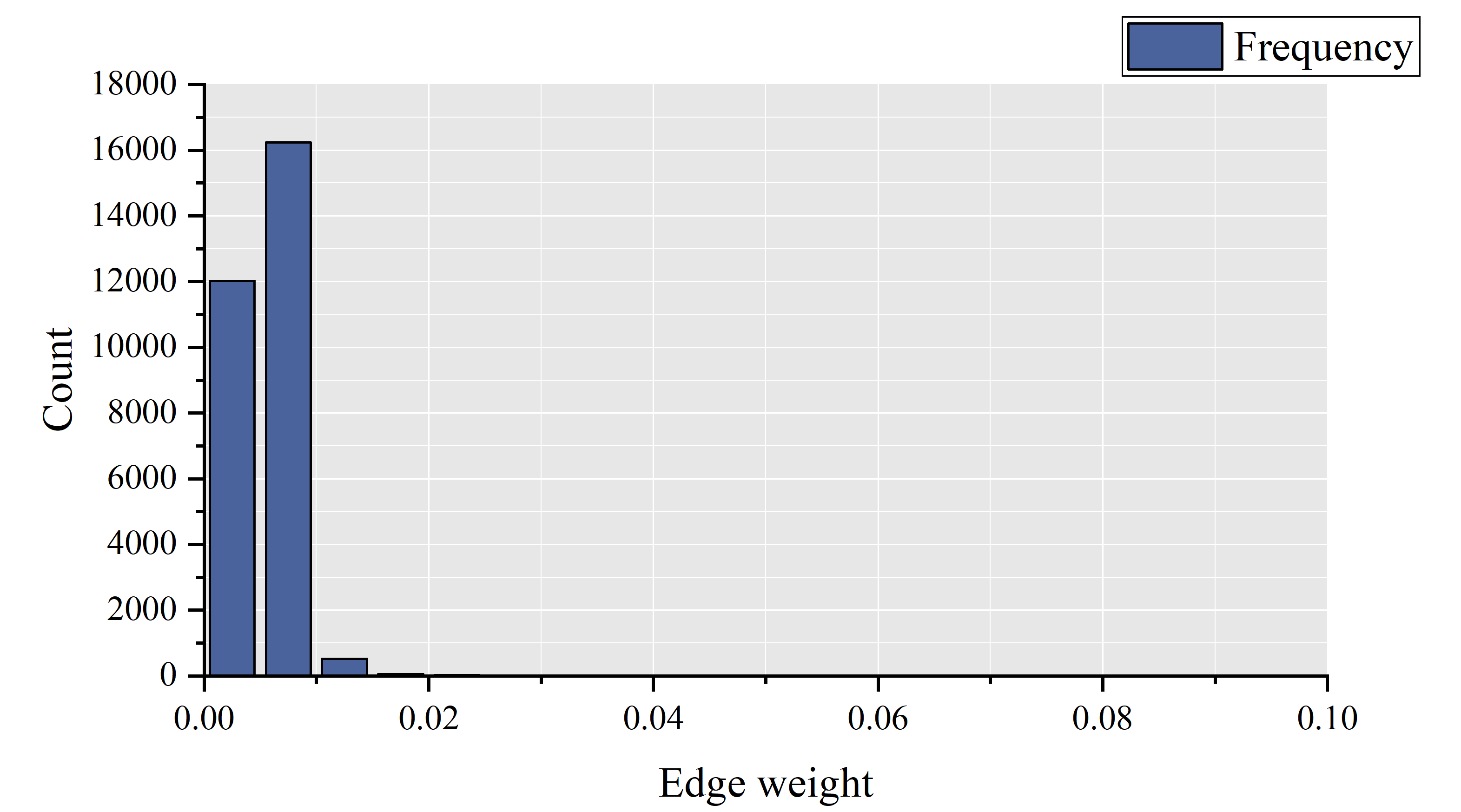}
    \caption{Histogram of edge weights of the spatial graph from an ASTGNN trained on the PeMSD8 dataset.}
    \label{fig:longtail}
\end{figure}

In this work, we ask the following question: \textbf{whether and to what extent can we localise spatial-temporal graph models?}
We limit our investigations to ASTGNNs, as they represent the state-of-the-art spatial-temporal graph model architecture.
ASTGNNs usually model spatial dependencies with adaptive graph convolutional layers.
The spatial dependencies are captured by learning the graph adjacency matrices.
Therefore, the localisation of an ASTGNN is achieved via sparsifying the adjacency matrices, which can also be understood as pruning edges in the spatial graph.
Note that the localisation of an ASTGNN refers to the sparsification of only the adjacency matrices capturing the spatial dependencies.
It is crucial not to be confused with sparsifying other weight matrices, such as those used in the temporal modules, often seen in GNN sparsification with pre-defined graph architectures \cite{bib:PAKDD20:Li, bib:ICML20:Zheng, bib:ICML21:Chen2, bib:AAAI22:You}.
An initial clue which indicates that the localisation of ASTGNNs could be feasible, probably even to an extreme extent, is shown in \figref{fig:longtail}).
Here we show the distribution of the elements in the spatial graph's adjacency matrix from an ASTGNN trained on the PeMSD8 dataset \cite{bib:others01:Chen}.
It is obvious that most edges in the spatial graph have weights close to zero.

We are interested in the localisation of spatial-temporal graph models for the following reasons:
\begin{itemize}
     \item \textit{A deeper understanding of the spatial and temporal dependencies in the data}.
     Although it is commonly accepted that both spatial and temporal dependencies are vital for inference, it is unclear whether and to what extent the information provided by these dependencies overlaps.
     If the localisation induces a marginal accuracy drop, then the information provided by the spatial dependencies is already largely included in the information contained in the temporal dependencies and, thus, unnecessary for inference.
     \item \textit{Resoruce-efficient ASTGNN designs}.
     ASTGNNs are notoriously computation heavy as the size of the spatial graph grows quadratically with the number of vertices, thus limiting their usage on large-scale data and applications. 
     The localisation of ASTGNNs may significantly reduce the resource requirement of these spatial-temporal graph models and enable new spatial-temporal applications.
     \item \textit{Distributed deployment of spatial-temporal graph models}.
     In many cases, the data to construct the spatial-temporal graph models are collected via distributed sensing systems \eg sensor networks. 
     However, making predictions of each vertex using these models requires the history of other vertices, thereby involving data exchange between sensor nodes.
     Localisation of spatial-temporal graph models may enable individual sensor nodes to make predictions autonomously without communicating with each other, which saves bandwidth and protects privacy in a distributed system.
\end{itemize}

\begin{figure*}[t]
    \centering
    \includegraphics[width=1\linewidth]{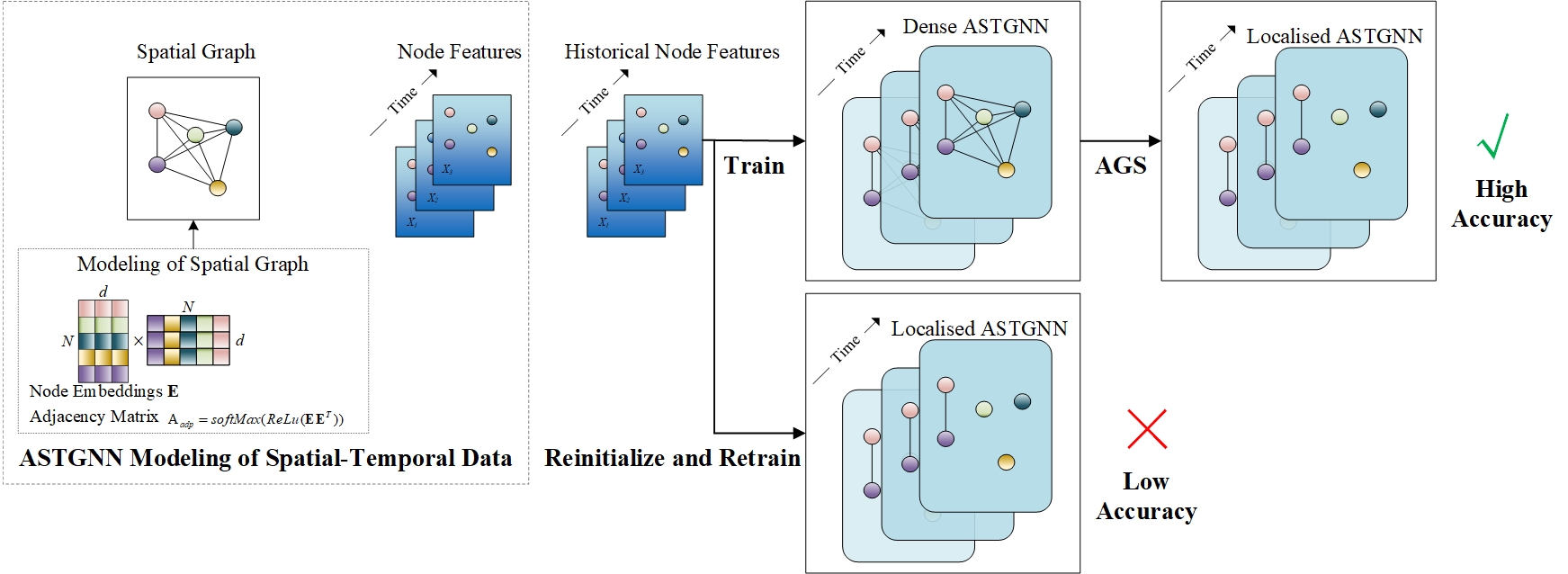}
    \caption{An overview of our experiments and observations. We train ASTGNNs on spatial-temporal datasets, achieving baseline accuracies. Then we localise the ASTGNNs with the proposed algorithm \sysname, achieving accuracies comparable to the dense graph baselines. Finally, we reinitialise the localised ASTGNNs and retrain them on the same datasets, resulting in considerably and consistently deteriorated accuracies.}
    \label{fig:overview}
\end{figure*}

We explore the localisation of ASTGNNs via Adaptive Graph Sparsification (\sysname), a novel algorithm dedicated to the sparsification of adjacency matrices in ASTGNNs.
The core of \sysname is a differentiable approximation of the $L_{0}$-regularization of a mask matrix, which allows the back-propagation to go past the regularizer and thus enables a progressive sparsification while training.
We apply \sysname to two representative ASTGNN architectures and nine different spatial-temporal datasets. 
The experiment results are surprising.
\textit{(i)} 
The spatial adjacency matrices can be sparsified to over $99.5\%$ without deterioration in test accuracy on all datasets.
\textit{(ii)} 
Even fully localised ASTGNNs, which effectively degenerate to purely temporal models, can still provide decent test accuracy (no deterioration on most tested datasets while only minor accuracy drops on the rest).
\textit{(iii)} 
When we reinitialise the weights of the localised ASTGNNs and retrain them on the spatial-temporal datasets, we cannot reinstate the same inference accuracy.
\figref{fig:overview} summarises our experiments and observations.

Our empirical study implies two hypotheses.
\textit{(i)} 
In the tested spatial-temporal datasets, the information provided by the spatial dependencies is primarily included in the information provided by the temporal dependencies. Therefore, \textbf{the spatial dependencies can be safely ignored for inference} without a noteworthy loss of accuracy.
\textit{(ii)} 
Although the information contained in the spatial and temporal dependencies overlaps, such overlapping provides the vital redundancy necessary for properly training a spatial-temporal graph model. 
Thus, \textbf{the spatial dependencies cannot be ignored during training}.

Our main contributions are summarised as follows:
\begin{itemize}
    \item
    To the best of our knowledge, this is the first study on the localisation of spatial-temporal graph models.
    We surprisingly observed that spatial dependencies could be largely ignored during inference without losing accuracy.
    Extensive experiments on common spatial-temporal datasets and representative ASTGNN architectures demonstrated that only a few edges (less than $0.5\%$ on all tested datasets) are required to maintain the inference accuracy.
    More surprisingly, when the spatial dependencies are completely ignored, \ie the ASTGNNs are fully localised, they can still maintain a decent inference accuracy (no deterioration on most tested datasets, minor drops on the rest).
    \item
    With further investigations, we suggest the hypothesis that, although spatial dependencies can be primarily ignored during inference, they can drastically improve training effectiveness.
    This is supported by the observation that, if we reinitialise all parameters in the sparsified ASTGNNs and retrain them with the same data, the retrained networks yield considerably and consistently worse accuracy.
    \item 
    To enable the localisation of ASTGNNs, we propose Adaptive Graph Sparsification (\sysname), a novel graph sparsification algorithm dedicated to ASTGNNs. The core of \sysname is a differentiable approximation of the $L_{0}$-regularization of a mask matrix, which allows the back-propagation to go past the regularizer and thus enables a progressive sparsification while training.
\end{itemize}

\label{sec:introduction}

\section{Related Work}
\label{sec:related}
Our work is relevant to the following threads of research.
\subsection{Spatial-Temporal Graph Neural Networks}
Spatial-temporal graph neural networks (STGNNs) play an essential role in spatial-temporal data analysis for their ability to learn hidden patterns of spatial irregular signals varying across time \cite{bib:TNNLS20:Wu}. 
These models often combine graph convolutional networks and recurrent neural networks. 
For example, Graph Convolutional Recurrent Network (GCRN) \cite{bib:ICONIP18:Seo} combines an LSTM with ChebNet. Diffusion Convolutional Recurrent Neural Network \cite{bib:ICLR18:Li} incorporates a proposed diffusion graph convolutional layer into GRU in an encoder-decoder manner to make multi-step predictions.
Alternatively, CNN-based models can represent the temporal relations in spatial-temporal data in a non-recursive manner. 
For instance, CGCN \cite{bib:IJCAI18:Yu} combines 1D convolutional layers with GCN layers.
ST-GCN \cite{bib:AAAI18:Yan} composes a spatial-temporal model for skeleton-based action recognition using a 1D convolutional layer and a Partition Graph Convolution (PGC) layer.
More recent proposals such as ASTGCN \cite{bib:AAAI19:Guo}, STG2Seq \cite{bib:IJCAI:bai}, and LSGCN \cite{bib:IJCAI20:Huang} further employ attention mechanisms to model dynamic spatial dependencies and temporal dependencies. In addition, some researchers consider the out-of-distribution generalisation of STGNN, and propose a domain generalisation framework based on hypernetworks to solve this problem\cite{DBLP:conf/icpads/DuanHZTR22}. However, these models adopt a predefined graph structure, which may not reflect the complete spatial dependency. 

To capture the dynamics in graph structures of spatial-temporal data, an emerging trend is to utilize adaptive spatial-temporal graph neural networks (ASTGNNs).
Graph WaveNet \cite{bib:IJCAI19:Wu} proposes an AGCN layer to learn a normalized adaptive adjacency matrix without a pre-defined graph. 
ASTGAT introduces a network generator model that generates an adaptive discrete graph with the Gumbel-Softmax technique\cite{kong2022adaptive}. The network generator can adaptively infer the hidden correlations from data.  AGCRN \cite{bib:NIPS20:Bai} designs a Node Adaptive Parameter Learning enhanced AGCN (NAPL-AGCN) to learn node-specific patterns. 
Due to its start-of-the-art performance, NAPL-AGCN has been integrated into various recent models such as Z-GCNETs \cite{bib:ICML21:Chen}, STG-NCDE \cite{bib:AAAI22:Choi}, and TAMP-S2GCNets \cite{bib:ICLR22:Chen}.

Despite the superior performance of ASTGNNs, they incur tremendous computation overhead, mainly because 
\textit{(i)} learning an adaptive adjacency matrix involves calculating the edge weight between each pair of nodes, and 
\textit{(ii)} the aggregation phase is computationally intensive.
We aim at efficient ASTGNN inference, particularly for large graphs.

\subsection{Graph Sparsification for GNNs}
With graphs rapidly growing, the training and inference cost of GNNs has become increasingly expensive. 
The prohibitive cost has motivated growing interest in graph sparsification.
The purpose of graph sparsification is to extract a small sub-graph from the original large one.
SGCN \cite{bib:PAKDD20:Li} is the first to investigate graph sparsification for GNNs, \ie pruning input graph edges, and learned an extra DNN surrogate. 
NeuralSparse \cite{bib:ICML20:Zheng} prunes task-irrelevant edges from downstream supervision signals to learn robust graph representation. 
More recent works such as UGS \cite{bib:ICML21:Chen2} and GBET \cite{bib:AAAI22:You} explore graph sparsification from the perspective of the winning tickets hypothesis.

The aforementioned works only explore graph sparsification for vanilla GNNs and non-temporal data with pre-defined graphs.
Our work differs by focusing on \textit{spatial-temporal} GNNs with \textit{adaptive} graph architectures.



\section{Preliminaries}
\label{subsec:std}
This section provides a quick review of the representative architectures of ASTGNNs.

\subsection{Spatial-Temporal Data as Graph Structure}
Following the conventions in spatial-temporal graph neural network research \cite{bib:TNNLS20:Wu, bib:ICONIP18:Seo, bib:ICLR18:Li, bib:IJCAI18:Yu, bib:AAAI18:Yan, bib:AAAI19:Guo, bib:IJCAI:bai, bib:IJCAI20:Huang, bib:IJCAI19:Wu, bib:NIPS20:Bai, bib:ICML21:Chen, bib:AAAI22:Choi, bib:ICLR22:Chen}, we represent the spatial-temporal data as a sequence of discrete frames $\mathcal{X}$ with $\mathcal{G}$ = $\{\mathcal{V}, \mathcal{E}\}$, where $\mathcal{X}=\{\boldsymbol{X}_1, \boldsymbol{X}_2, \ldots, \boldsymbol{X}_T\}$.
The graph $\mathcal{G}$ is also known as the spatial network, which consists of a set of nodes $\mathcal{V}$ and a set of edges $\mathcal{E}$. 
Let $\left|\mathcal{V}\right|=N$.
Then the edges are presented with an adjacency matrix $A_{t} \in \mathbb{R}^{N \times N}$, where $\boldsymbol{X}_t \in \mathbb{R}^{N \times C}$ is the node feature matrix with dimension $C$ at timestep $t$, for $t= 1, \ldots, T$.

Given graphs $\mathcal{G}$ and $\mathcal{T}$ historical observations $\mathcal{X}^{\mathcal{T}}$=$\{\boldsymbol{X}_{t-\mathcal{T}}$, $\ldots$, $\boldsymbol{X}_{t-1}\} \in \mathbb{R}^{\mathcal{T} \times N \times C}$, we aim to learn a function $\mathcal{F}$ which maps the historical observations into the future observations in the next $\mathcal{H}$ timesteps:
\begin{equation}
\left\{\boldsymbol{X}_t, \ldots, \boldsymbol{X}_{t+\mathcal{H}}\right\} = \mathcal{F}( \boldsymbol{X}_{t-\mathcal{T}}, \ldots, \boldsymbol{X}_{t-1}; \theta, \mathcal{G})
\end{equation}
where $\theta$ denotes all the learnable parameters. 

\subsection{Modeling the Spatial Network $\mathcal{G}$}
\label{subsec:gcn}
Since we focus on the role of spatial dependencies, we explain the state-of-the-art modelling of the spatial network in spatial-temporal graph neural networks (STGNNs).

The basic method to model the spatial network $\mathcal{G}$ at timestep $t$ with its node feature matrix $\boldsymbol{X}_t$ in representative STGNNs \cite{bib:IJCAI19:Wu,bib:IJCAI:bai,bib:IJCAI20:Huang} is Graph Convolution Networks (GCNs).
A one-layer GCN can be defined as:
\begin{equation}
Z_{t} =\sigma\left(\tilde{D}^{-\frac{1}{2}} \tilde{A}_{t} \tilde{D}^{-\frac{1}{2}} \boldsymbol{X}_t W\right)
\label{eq:gcn}
\end{equation}
where $\tilde{A} = A+I_{N}$ is the adjacency matrix of the graph with added self-connections.
$I_{N}$ is the identity matrix.
$\tilde{D}$ is the degree matrix.
$ W \in \mathbb{R}^{C \times F}$ is a trainable parameter matrix.
$\sigma(\cdot)$ is the activation function. 
$Z_{t} \in \mathbb{R}^{N \times F}$ is the output. 
All information regarding the input $\boldsymbol{X}_t$ at timestep $t$ is aggregated in $Z_{t}$.

A key improvement to model the spatial network is to adopt Adaptive Graph Convolution Networks (AGCNs) to capture the dynamics in the graph $\mathcal{G}$, which leads to the adaptive spatial-temporal graph neural networks (ASTGNNs) \cite{bib:IJCAI19:Wu,bib:NIPS20:Bai,bib:ICML21:Chen,bib:AAAI22:Choi,bib:ICLR22:Chen}.
In the following, we briefly explain Adaptive Graph Convolutional Recurrent Network (AGCRN) \cite{bib:NIPS20:Bai} and the extension with transformers, denoted as AGFormer, two representative ASTGNN models.
\begin{itemize}
    \item \textbf{AGCRN}.
    It enhances the GCN layer by combining the normalized self-adaptive adjacency matrix with a Node Adaptive Parameter Learning (NAPL), which is known as NAPL-AGCN.
    \begin{equation}
    \begin{gathered}
    \mathbf{A}_{a d p}={SoftMax}\left({ReLU}\left(\mathbf{E}\mathbf{E}^T\right)\right) \\
    Z_{t}=\sigma\left(\mathbf{A}_{a d p} \boldsymbol{X}_t \mathbf{E}W_{\mathcal{G}}\right)
    \end{gathered} \label{eq:self-adj}
    \end{equation}
    where $W_\mathcal{G} \in \mathbb{R}^{d \times C\times F}$ and $\mathbf{E}W_{\mathcal{G}} \in \mathbb{R}^{N \times C\times F}$.
    $\mathbf{A}_{a d p} \in \mathbb{R}^{N \times N}$ is the normalized self-adaptive adjacency matrix \cite{bib:IJCAI19:Wu}.
    $d$ is the embedding dimension for $d\ll N$. 
    Each row of $\mathbf{E}$ presents the embedding of the node. 
    During training, $\mathbf{E}$ is updated to learn the spatial dependencies among all nodes. 
    Instead of directly learning $W$ in \eqref{eq:gcn} shared by all nodes, NAPL-AGCN uses $\mathbf{E}W_{\mathcal{G}}$ to learn node-specific parameters.
    From the view of one node (\eg node $i$), $\mathbf{E}_{i}W_{\mathcal{G}}$ are the corresponding node-specific parameters according to its node embedding $\mathbf{E}_{i}$. 
    Finally, to capture both spatial and temporal dependencies, AGCRN integrates NAPL-AGCN and Gated Recurrent Units (GRU) by replacing the MLP layers in GRU with NAPL-AGCN.
    \item \textbf{AGFormer}.
    It extends AGCRN by modelling the temporal dependencies with transformers \cite{vaswani2017attention, li2019enhancing}. 
    A Transformer is a stack of transformer blocks. 
    One block contains a multi-head self-attention mechanism and a fully connected feed-forward network.
    We replace the MLP layers in the multi-head self-attention mechanism with NAPL-AGCN to construct a transformer-based ASTGNN model, which we call AGFormer.
\end{itemize}

Modelling the spatial network with NAPL-AGCN achieves state-of-the-art performances on multiple benchmarks, and it has been widely adopted in diverse ASTGNN variants \cite{bib:NIPS20:Bai, bib:ICML21:Chen, bib:AAAI22:Choi, bib:ICLR22:Chen}. 
However, NAPL-AGCN is much more inefficient than GCNs, since $\mathbf{A}_{a d p}$ is a matrix without zeros, while the adjacency matrix of a pre-defined graph in GCNs is far more sparse than $\mathbf{A}_{a d p}$.
This motivates us to explore the localisation of spatial-temporal graph models taking ASTGNNs as examples. 

\section{Adaptive Graph Sparsification}
\label{sec:method}
This section presents Adaptive Graph Sparsification (\sysname), a new algorithm dedicated to the sparsification of adjacency matrics in ASTGNNs.

\fakeparagraph{Formulation}
The NAPL-AGCN-based ASTGNNs with normalized self-adaptive adjacency matrix $\mathbf{A}_{a d p}$ can be trained using the following objective:
\begin{equation}\label{loss:1}
\mathcal{L}(\theta, \mathbf{A}_{a d p})=\frac{\sum_{\tau \in \mathrm{T}} \sum_{v \in \mathcal{V}}\left\|\boldsymbol{y}^{(\tau, v)}-\hat{\boldsymbol{y}}^{(\tau, v)}\right\|_1}{|\mathcal{V}| \times|\mathrm{T}|}
\end{equation}
where $\mathrm{T}$ is a training set, $\tau$ is a train sample, and $\boldsymbol{y}^{(\tau, v)}$ is the ground-truth of node $v$ in $\tau$. 

Given a pre-trained model $\mathcal{F}(\cdot;\theta,\mathbf{A}_{a d p})$, we introduce a mask $\mathbf{M}_{\mathbf{A}}$ to prune the adjacency matrix $\mathbf{A}_{a d p}$.
The shape of $\mathbf{M}_{\mathbf{A}}$ is identical to $\mathbf{A}_{a d p}$. 
Specifically, given $\mathcal{F}(\cdot;\theta,\mathbf{A}_{a d p})$, we obtain $\mathbf{M}_\mathbf{A}$ by optimizing the following objective:
\begin{equation}\label{loss:2}
\begin{gathered}
\mathcal{L}_{AGS}=\mathcal{L}(\theta, \mathbf{A}_{a d p}\odot \mathbf{M}_\mathbf{A})+
\lambda\left\|\mathbf{M}_{\mathbf{A}}\right\|_0,\\
\left\|\mathbf{M}_{\mathbf{A}}\right\|_0=\sum_{i=1}^N \sum_{j=1}^N m_{(i, j)}, m_{(i,j)} \in \left\{0,1\right\}
\end{gathered}
\end{equation}
where $\odot$ is the element-wise product, $m_{(i, j)}$ corresponds to binary ``gate'' that indicates whether an edge is pruned, and $\lambda$ is a weighting factor for $L_{0}$-regularization of $\mathbf{M}_\mathbf{A}$.

\fakeparagraph{Sparsification Algorithm}
An intuitive way to get $\mathbf{M}_\mathbf{A}$ is to initialize a trainable weight matrix $\mathbf{U} \in \mathbb{R}^{N \times N}$ and map the entry $u_{(i, j)} \in \mathbf{U}$ into binary ``gates'' using a Bernoulli distribution: $m_{(i, j)}=\mathcal{B}\left(u_{(i, j)}\right)$, where $\mathcal{B}$ is a Bernoulli distribution. 
However, directly introducing $\mathbf{U}$ to model $\mathcal{F}(\cdot;\theta,\mathbf{A}_{a d p})$ has two problems. 
\begin{itemize}
    \item It may be unscalable for large-scale graphs.
    \item The $L_{0}$ sparsity penalty is non-differentiable.
\end{itemize}


For the scalability issue, we adopt a node adaptive weight learning technique to reduce the computation cost by simply generating $\mathbf{U}$ with the node embedding $\mathbf{E}$:
\begin{equation}\label{eq:V}
\mathbf{U}=\mathbf{E} W_{\mathbf{E}}
\end{equation}
where $W_{\mathbf{E}} \in \mathbb{R}^{d \times N}$ is a trainable matrix. 

For the non-differentiable issue, we introduce the hard concrete distribution instead of the Bernoulli distribution \cite{bib:ICLR18:Chris}, which is a contiguous relaxation of a discrete distribution and can approximate binary values.

Accordingly, the computation of binary ``gates'' $m_(i,j)$ can be formulated as:
\begin{equation}
\begin{gathered}
z \sim \mathcal{U}(0,1), s_{(i, j)}= Sig\left(\log z-\log (1-z)+\log \left(u_{(i, j)}\right)\right) / \beta \\
\bar{s}_{(i, j)}=s_{(i, j)}(\zeta-\gamma)+\gamma, m_{(i, j)}=\min \left(1, \max \left(0, \bar{s}_{(i, j)}\right)\right)
\end{gathered}
\end{equation}
where $z$ is a uniform distribution, $Sig$ is a sigmoid function, $\beta$ is a temperature value and $(\zeta-\gamma)$ is the interval with $\zeta<0$ and $\gamma> 1$. 
We set $\zeta=-0.1$ and $\gamma=1.1$ in practice. 
Then, $\mathbf{M}_{\mathbf{A}}$ is applied to prune the lowest-magnitude entries in $\mathbf{A}_{a d p}$, \wrt pre-defined ratios $p_g$. 

\begin{algorithm}[t]
    \renewcommand{\algorithmicrequire}{\textbf{Input:}}
	\renewcommand{\algorithmicensure}{\textbf{Output:}}
  \caption{ Adaptive Graph Sparsification (AGS)}
  \label{alg:AGS}
  \begin{algorithmic}[1]
    \REQUIRE
      $\mathcal{X}$: input data,
      $\mathcal{F}\left(\cdot; \theta, \mathbf{A}_{a d p}\right)$: Spatial-temporal GNN \\ with initialization self-adaptive adjacency matrix $\mathbf{A}_{a d p}$, \\ $\mathrm{N}_1$: number of pre-training iterations, $\mathrm{N}_2$: number of sparsification iterations,
      $s_g$: pre-defined sparsity level for graph.
      
    \ENSURE $\mathcal{F}\left(\cdot; \theta, \mathbf{A}_{a d p}\odot \mathbf{M}_\mathbf{A}\right)$
    \WHILE{iteration $i<\mathrm{N}_1$}
    \STATE Forward to compute the loss in \equref{loss:1}.
    \STATE Back-propagate to update $\theta$ and $\mathbf{A}_{a d p}$ .
    \ENDWHILE
    \STATE Obtain pre-trained $\mathcal{F}\left(\cdot; \theta, \mathbf{A}_{a d p}\right)$ .
    \STATE Sort entries in $\mathbf{A}_{a d p}$ by magnitude in an ascending order \\then obtain list $K = \left\{k_{i}\right\}_{i=1}^{N \times N}$.
    \WHILE{iteration $i<\mathrm{N}_{2}$ and $1-\frac{\left\|\mathbf{M}_{\mathbf{A}}\right\|_0}{\left\|\mathbf{A}_{a d p}\right\|_0}<s_{g}$}
    \STATE set $\mathbf{M}_\mathbf{A}^{(i,j)}=1$ if $\mathbf{A}_{a d p}^{(i,j)} \notin K_{:N^{2}s_{g}}$.
    \STATE Forward to compute the loss in \equref{loss:2}.
    \STATE Back-propagate to update $\theta$ and $\mathbf{A}_{a d p} \odot \mathbf{M}_\mathbf{A}$ .
    \ENDWHILE
  \end{algorithmic}
\end{algorithm}

\algoref{alg:AGS} outlines the procedure of \sysname.
The pruning begins after the network is pre-trained (Line 5). 
First, we sort edges in the adjacency matrix by magnitude in ascending order (Line 6).  
Then we perform pruning iteratively (Line 7). 
The top $p_{g}$ of the edges are removed, and the rest are retained. 
We identify the remaining edge by setting its corresponding entry in $\mathbf{M}_{\mathbf{A}}$ to 1 in each iteration (Line 8).
The edges to be pruned are removed using \equref{loss:2} (Line 9).

\fakeparagraph{Discussions}
We make two notes on our \sysname algorithm.
\begin{itemize}
    \item 
    \sysname differs from prior magnitude-based pruning methods designed for graphs \cite{bib:ICML21:Chen2, bib:AAAI22:You} in that
    \textit{(i)} it sparsifies  \textit{adaptive graphs} in spatial-temporal GNNs rather than vanilla GNNs and non-temporal data with pre-defined graphs; and 
    \textit{(ii)} it does not require iterative retraining on the pruned graphs to restore model accuracy.
    \item
    Pruning the adjacency matrix with \sysname notably reduces the complexity of ASTGNNs for inference.
    The inference time complexity of unpruned NAPL-AGCN layers is $\mathcal{O} (N^2d+ \\LN\mathcal{T} F^2+L\mathcal{T} \left\|\mathrm{A}_{a d p} \right\|_0  F+ NdF)$. 
    After sparsification, the inference time complexity of NAPL-AGCN layers is $\mathcal{O} (N^2d + LN\mathcal{T} F^2+L\mathcal{T} \left\|\mathrm{A}_{a d p} \odot \mathbf{M}_{\mathrm{A}} \right\|_0 F+NdF)$, where $N^2 d$ is the time complexity of computing adaptive adjacency matrix, $d$
is the embedding dimension,$N$ is the number of nodes, $\left\|A_{a d p} \odot M_A\right\|_0$ is the number of remaining edges, $F$ is the representation dimension of node features, $\mathcal{T}$ is the length of input, $L$ is the number of layers. 
\end{itemize}


\begin{table}[!t]
\centering
\caption{Summary of datasets used in experiments.}\label{tab:datasets}
\begin{tabular}{lcc}
    \toprule
     Datasets&\#Nodes& Range\\
    \midrule
     PeMSD3&358&09/01/2018 - 30/11/2018\\ 
     PeMSD4&307&01/01/2018 - 28/02/2018\\
     PeMSD7&883&01/07/2017 - 31/08/2017\\
     PeMSD8&170&01/07/2016 - 31/08/2016\\
     \midrule
     Bytom&100&27/07/2017 - 07/05/2018\\
     Decentral&100&14/10/2017 - 07/05/2018\\
     Golem&100&18/02/2017 - 07/05/2018\\
     \midrule
     CA&55&01/02/2020 - 31/12/2020\\
     TX&251&01/02/2020 - 31/12/2020\\
     \bottomrule
\end{tabular}
\end{table}

\section{Experiments}
\label{sec:evaluation}
To answer the question of whether and to what extent we can localise a spatial-temporal graph model, we conducted extensive experiments explained in this section.
\subsection{Neural Network Architecture}
We evaluate the performance of \sysname over two representative NAPL-AGCN-based ASTGNN architectures: \textbf{AGCRN} \cite{bib:NIPS20:Bai} and its extension \textbf{AGFormer}.
AGCRN is a state-of-the-art ASTGNN architecture combining AGCN and RNN layers.
AGCN layers are used to capture the spatial dependencies, while RNN layers are there to model the temporal dependencies.
AGFormer, on the other hand, can be regarded as an alternative version of the AGCRN, in which the RNN layers are substituted by Transformer layers.
We intentionally chose these two ASTGNN architectures sharing the same spatial module but using different temporal modules, to show that both the effectiveness of \sysname and our observations on the learned spatial dependencies are orthogonal to the temporal modules involved.

\subsection{Datasets and Configurations}
The localisation of ASTGNNs is evaluated on nine real-world spatial-temporal datasets from three application domains: transportation, blockchain and biosurveillance. \tabref{tab:datasets} summarizes the specifications of the datasets used in our experiments. The detailed datasets and configurations are provided in \appref{app:datasets}. 
The details on tuning hyperparameters are provided in \appref{app:detail}.



\begin{figure*}[t]
    \centering
    \includegraphics[width=1\linewidth]{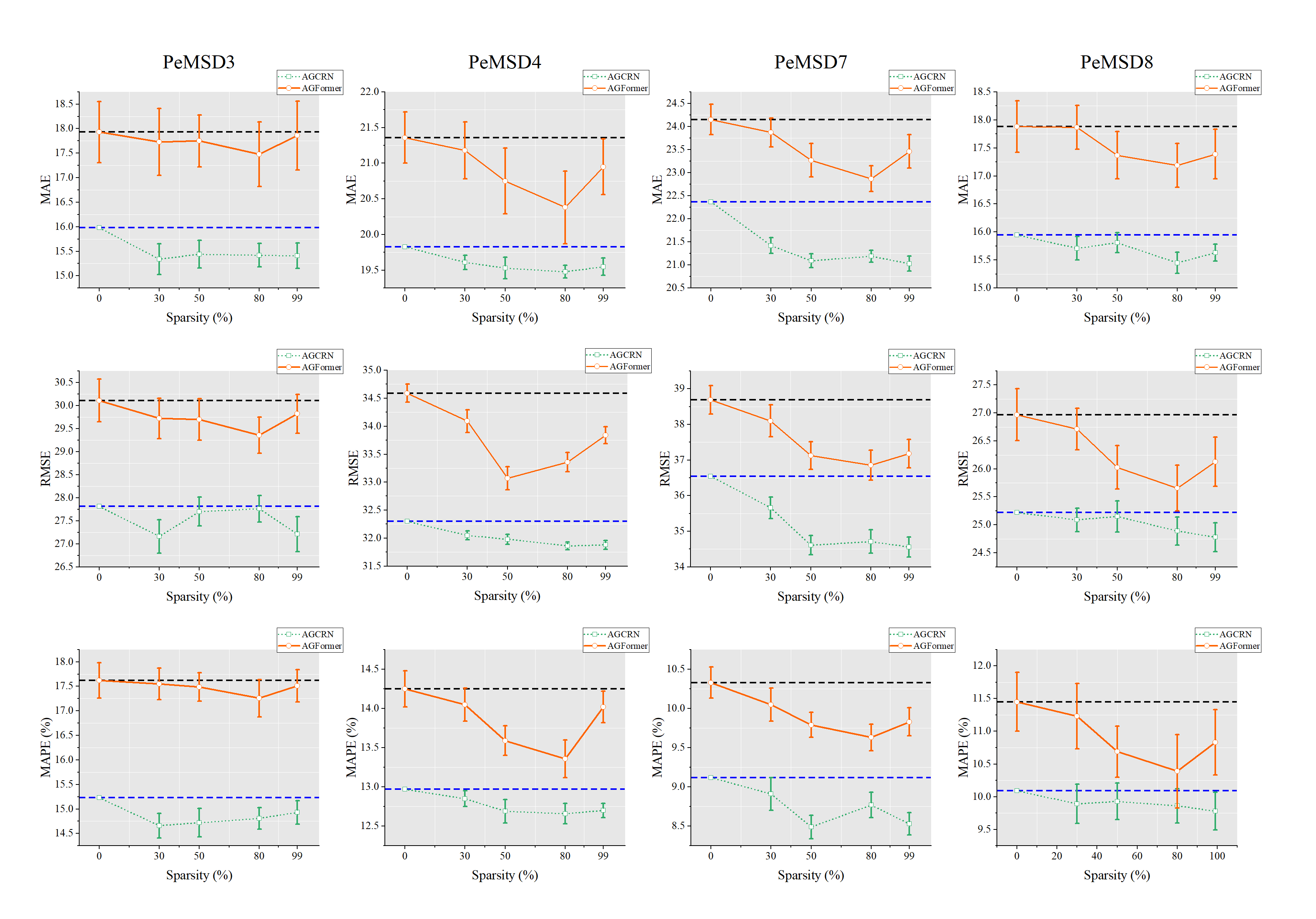}
    \caption{Test accuracies of original and localised (up to 99\%) AGCRNs and AGFormers, tested on transportation datasets (PeMSD3, PeMSD4, PeMSD7 and PeMSD8).  Horizontal dash lines represent the baselines of non-localised AGCRNs and AGFormers.}
    \label{fig:pemsd}
\end{figure*}

\begin{figure}[ht]
    \centering
    \includegraphics[width=1\linewidth]{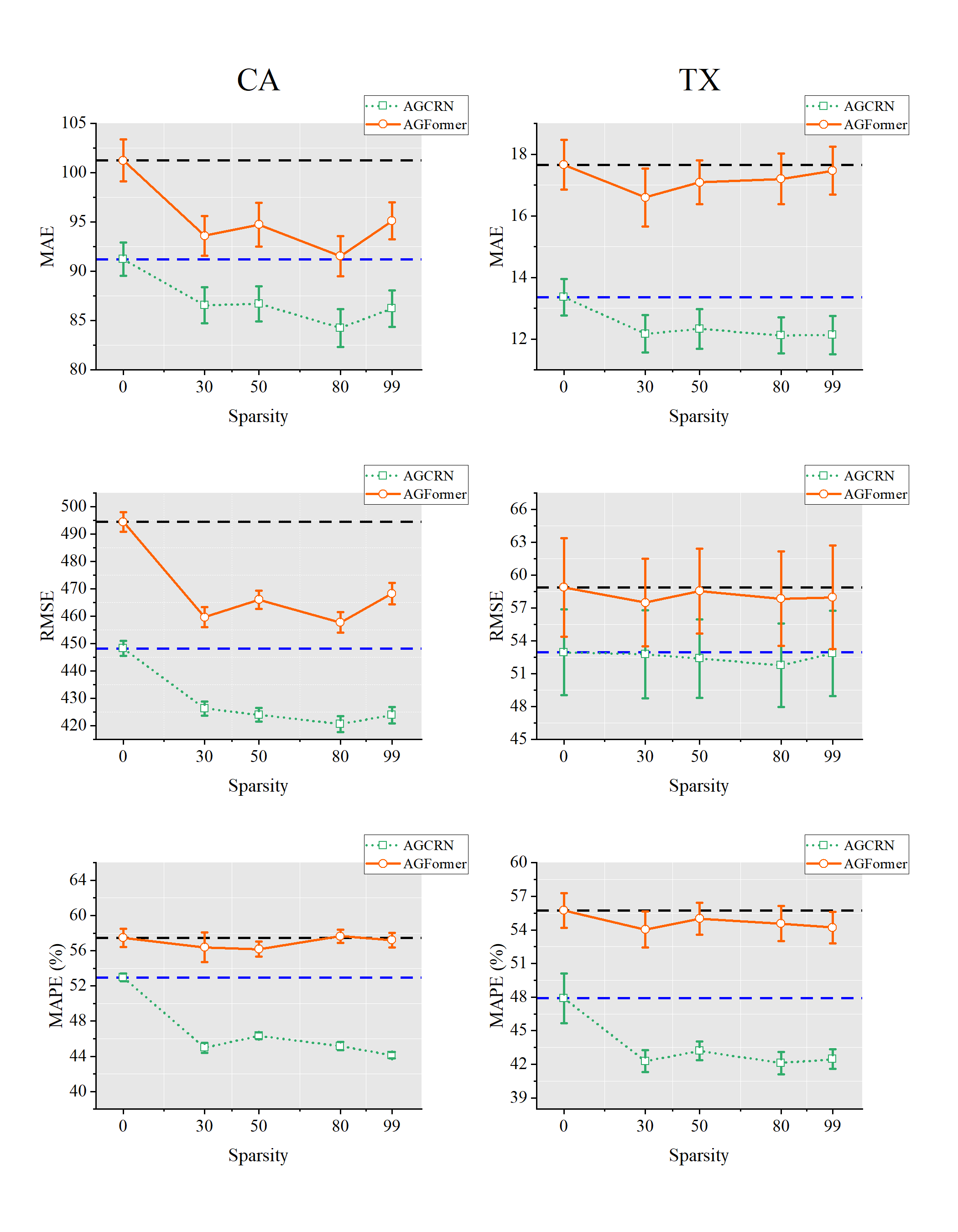}
    \caption{Test accuracies of original and localised (up to 99\%) AGCRNs and AGFormers, tested on biosurveillance datasets (CA and TX).  Horizontal dash lines represent the baselines of non-localised AGCRNs and AGFormers.}
    \label{fig:covid}
\end{figure}

\begin{figure*}[ht]
    \centering
    \includegraphics[width=1\linewidth]{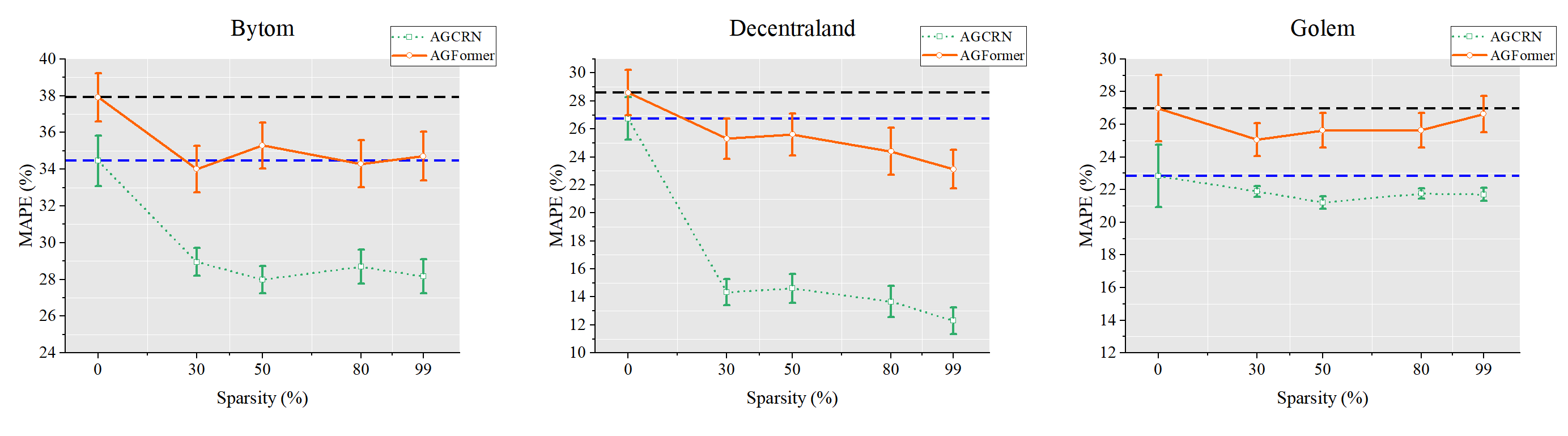}
    \caption{Test accuracies of original and localised (up to 99\%) AGCRNs and AGFormers, tested on blockchain datasets (Bytom, Decentraland and Golem).  Horizontal dash lines represent the baselines of non-localised AGCRNs and AGFormers.}
    \label{fig:ether}
\end{figure*}

\begin{figure*}[ht]

    \centering
    \includegraphics[width=1\linewidth]{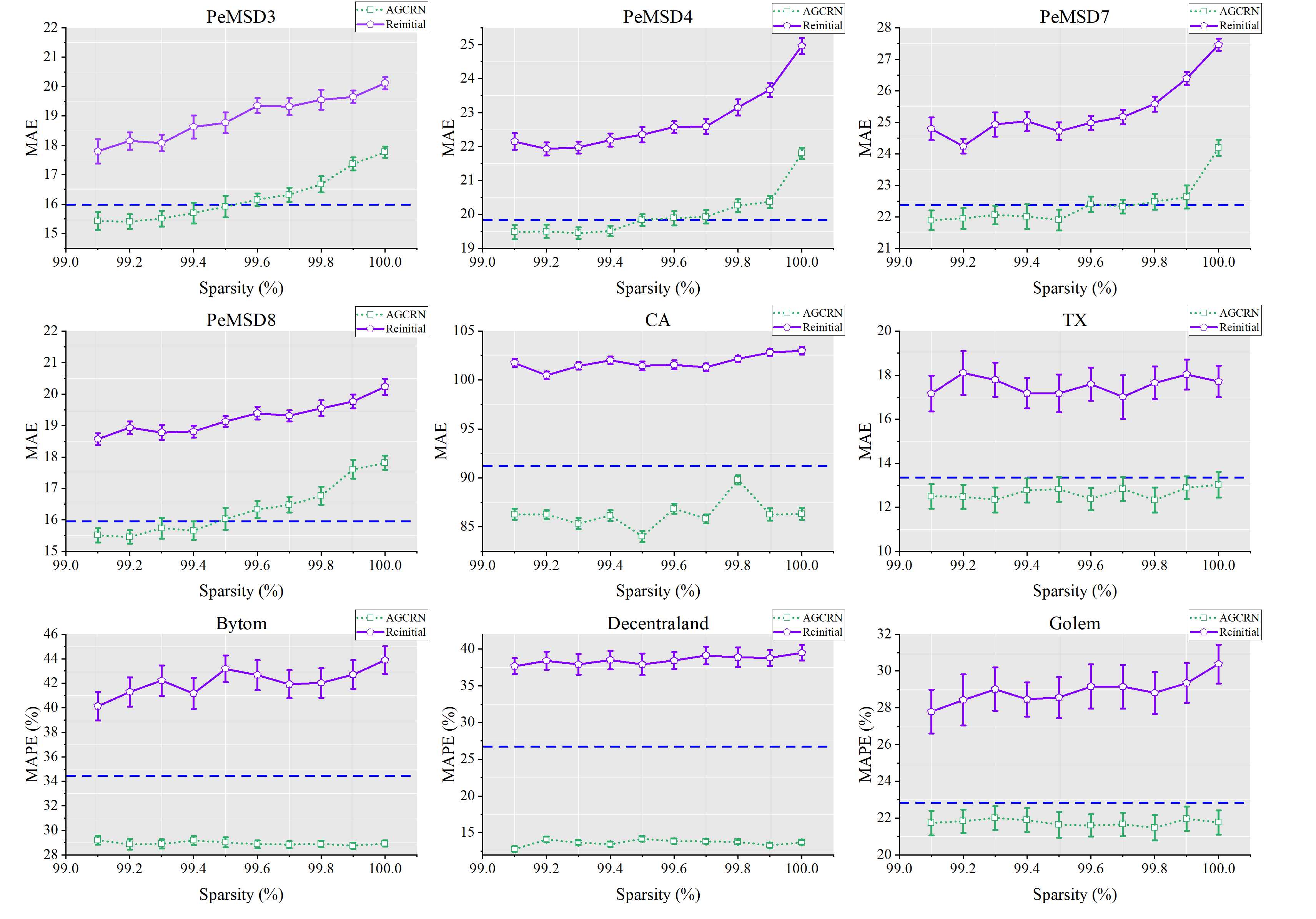}
    \caption{Test accuracies of localised (99.1\% to 100\%) AGCRNs and their reinitialised\&retrained counterparts, tested on all datasets. Horizontal dash lines represent the baselines of non-localised AGCRNs.}
    \label{fig:rewind}
\end{figure*}

\subsection{Main Experimental Results}
\label{sec:main_results}
Our major experiments are illustrated in \figref{fig:overview}.
We first train AGCRNs and AGFormers on the nine spatial-temporal datasets, achieving baseline accuracies. 
Then we conduct localisation of these well-trained AGCRNs and AGFormers using \sysname.
Finally, we reinitialise all weights in the localised AGCRNs and retrain them on the original datasets with the same training settings.

The experiment results are organised as follows:
\begin{itemize}
    \item The test accuracies of the non-localised AGCRNs and AGFormers and those with localisation degrees until $99\%$ are collected in \figref{fig:pemsd} (on transportation datasets), \figref{fig:covid} (on biosurveillance datasets) and \figref{fig:ether} (blockchain datasets). 
    These figures contain the test accuracies at graph sparsity of 0\%, 30\%,  50\%,  80\% and  99\%.
    \item The test accuracies of AGCRNs with localisation degrees between 99.1\% and 100\% are illustrated in \figref{fig:rewind}, shown in dotted green curves.
    \item The test accuracies of localised AGCRNs that are reinitialised and retrained are also collected in \figref{fig:rewind}, shown in solid purple curves.
\end{itemize}
Error bars in these figures show the standard deviation of five runs. 
We made the following observations from these results:
\begin{itemize}
    \item \textbf{The localisation of AGCRNs and AGFormers is possible.}
    Applying \sysname on AGCRNs and AGFormers and localising them to a localisation degree of 99\% incurs no performance degradation across all datasets.
    On the contrary, in many experiments, the test accuracy keeps improving until 99\%-localisation.
    Further localisation of AGCRNs up to 99.5\% still induces no accuracy drop against the non-localised baselines.
    
    \item \textbf{Full localisation of AGCRNs is still practical.}
    Even when we fully localise the AGCRNs, which in effect turn them into independent RNNs ignoring all spatial dependencies, they can still provide decent test accuracies.
    As shown in \figref{fig:rewind}, on transportation datasets (PeMSD3, PeMSD4, PeMSD7 and PeMSD8), only minor drops are observed.
    On blockchain datasets (Bytom, Decentraland and Golem) and biosurveillance datasets (CA\&TX), we can observe that the test accuracy is no worse at 100\% sparsity compared with the non-localised baselines.

    \item \textbf{Localised AGCRNs cannot be relearned without the dense spatial graphs.}
    As shown in \figref{fig:rewind}, when we reinitialise the partially or fully localised AGCRNs and then retrain them on the nine datasets, we can observe a consistent and considerable drop in inference accuracy.
\end{itemize}

Upon these observations, we suggest the following hypotheses:

\begin{itemize}
    \item 
    In many spatial-temporal datasets, the information provided by the spatial dependencies is primarily included in the information provided by the temporal dependencies. 
    Therefore, the spatial dependencies can be safely ignored for inference without a noteworthy loss of accuracy.

    \item
    Although the information contained in the spatial and temporal dependencies overlaps, such overlapping provides the vital redundancy necessary for properly training a spatial-temporal graph model. 
    Thus, the spatial dependencies cannot be ignored during training.
\end{itemize}

\begin{table*}[htbp]
\caption{Computation cost during inference on original and 99\%-localised AGCRNs and AGformsers. The amount of computation is measured in MFLOPs, and acceleration factors are calculated in the round brackets.}
\small
\setlength{\tabcolsep}{0.8mm}{
\begin{tabular}{lccccccccc}
\Xhline{1pt}
\multirow{2}{*}{Methods} & \multicolumn{9}{c}{Computation Cost for Inference (MFLOPs)}                                              \\ \cline{2-10} 
    & PeMSD3 & PeMSD4 &PeMSD7  & PeMSD8 & Decentraland & Bytom & Golem & CA     & TX     \\ 
\hline
Original AGCRN                    & 400.26 & 188.59 & 1131.41 & 153.06 & 8.58         & 8.58  & 8.58  & 161.42 & 850.29 \\
Original AGFormer                 & 122.01 & 99.45  & 453.89  & 47.39 & 15.02      & 15.02 & 15.02 & 21.50  & 266.97 \\
\hline
\textbf{Localised AGCRN}    & 253.33($\textcolor{green}{\uparrow}1.6\times$) & 80.55($\textcolor{green}{\uparrow}2.3\times$) & 237.56($\textcolor{green}{\uparrow}4.8\times$) & 119.93($\textcolor{green}{\uparrow}1.3\times$) & 2.82($\textcolor{green}{\uparrow}3.0\times$)   & 2.82($\textcolor{green}{\uparrow}3.0\times$)   & 2.82($\textcolor{green}{\uparrow}3.0\times$)   & 145.50($\textcolor{green}{\uparrow}1.1\times$) & 706.21($\textcolor{green}{\uparrow}1.2\times$) \\
\textbf{Localised AGFormer} & 80.13($\textcolor{green}{\uparrow}1.5\times$)  & 68.64($\textcolor{green}{\uparrow}1.4\times$)  & 199.17($\textcolor{green}{\uparrow}2.3\times$) & 37.95($\textcolor{green}{\uparrow}1.3\times$) & 11.75($\textcolor{green}{\uparrow}1.3\times$) & 1.75($\textcolor{green}{\uparrow}1.3\times$) & 11.75($\textcolor{green}{\uparrow}1.3\times$) & 19.56($\textcolor{green}{\uparrow}1.1\times$)  & 226.43($\textcolor{green}{\uparrow}1.1\times$) \\ 
\Xhline{1pt}
\end{tabular}
}
\label{tab:flops}
\end{table*}

\begin{table*}[htbp]
\caption{Performance of 99\%-localised AGCRNs compared with other non-localised ASTGNN architectures on transportation datasets.}
\footnotesize
\begin{tabular}{lllllllllllllllll}
\Xhline{1pt}
\multirow{2}{*}{Methods} &
  Datasets &
  \multicolumn{3}{c}{PeMSD3} &
  \multicolumn{3}{c}{PeMSD4} &
  \multicolumn{3}{c}{PeMSD7} &
  \multicolumn{3}{c}{PeMSD8} &
  \multicolumn{3}{c}{Average}\\ 
  \cline{2-17} 
             & Metrics           & MAE   & RMSE  & MAPE    & MAE   & RMSE  & MAPE    & MAE   & RMSE  & MAPE    & MAE   & RMSE  & MAPE    & MAE   & RMSE  & MAPE\\ 
             \hline
\multicolumn{2}{l}{AGCRN}        & 15.98 & 28.25 & 15.23\% & 19.83 & 32.30 & 12.97\% & 22.37 & 36.55 & 9.12\% & 15.95 & 25.22 & 10.09\% &18.53 &30.58 &11.85\% \\
\multicolumn{2}{l}{Z-GCNETs}     & 16.64 & 28.15 & 16.39\% & 19.50 & 31.61 & 12.78\% & 21.77 & 35.17 & 9.25\% & 15.76 & 25.11 & 10.01\% &18.42 & 30.01 & 12.11\% \\
\multicolumn{2}{l}{STG-NCDE} &
  15.57 &
  \textbf{27.09} &
  15.06\% &
  \textbf{19.21} &
  \textbf{31.09} &
  12.76\% &
  \textbf{20.53} &
  \textbf{33.84} &
  8.80\% &
  \textbf{15.45} &
  24.81 &
  9.92\% & \textbf{17.69} & \textbf{29.21}&11.64\% \\
\multicolumn{2}{l}{TAMP-S2GCNets} & 16.03 & 28.28 & 15.37\% & 19.58 & 31.64 & 13.22\% & 22.16 & 36.24 & 9.20\% & 16.17 & 25.75 & 10.18\% & 18.49 & 30.48 & 11.99\%\\
\hline
\multicolumn{2}{l}{\textbf{Localised AGCRN}} &
  \textbf{15.41} &
  27.21 &
  \textbf{14.93\%} &
  19.55 &
  31.88 &
  \textbf{12.70\%} &
  21.03 &
  34.56 &
  \textbf{8.53\%} &
  15.63 &
  \textbf{24.78} &
  \textbf{9.78\%} &17.91&29.61 &\textbf{11.49}\% \\ 
\Xhline{1pt}
\end{tabular}
\label{tab:traffic}
\end{table*}

\begin{table*}[t]
\caption{Performance of 99\%-localised AGCRNs compared with other non-localised ASTGNN architectures on biosurveillance datasets.}
\begin{tabular}{llllllll}
\Xhline{1pt}
\multirow{2}{*}{Methods} & Datasets & \multicolumn{3}{c}{CA}          & \multicolumn{3}{c}{TX}    \\ 
\cline{2-8} 
                         & Metrics  & MAE        & RMSE        & MAPE & MAE   & RMSE       & MAPE \\ 
                         \hline
\multicolumn{2}{l}{AGCRN}           & 91.23±1.69 & 448.27±2.78 & 53.67±0.42     & 13.36±0.58 & 52.96±3.92 &47.89\% ±2.22     \\
\multicolumn{2}{l}{TAMP-S2GCNets}    &  \textbf{76.53±0.87}       & \textbf{371.60±2.68} & 92.90±1.57     & \textbf{11.29±1.05}      & \textbf{48.21±3.17} & 52.34\% ±3.87    \\ 
\hline
\multicolumn{2}{l}{\textbf{Localised AGCRN}}      & 86.22±1.43           &   423.91±3.01          & \textbf{44.12±0.35}     &       12.31±0.59&   52.88 ±3.73    & \textbf{42.47\%±1.41}    \\ 
\Xhline{1pt}
\end{tabular}
\label{tab:covid}
\end{table*}

\begin{table}[htbp]
\small
\caption{Performance of 99\%-localised AGCRNs com-pared with other non-localised ASTGNN architectures on blockchain datasets.}
\begin{tabular}{llll}
\Xhline{1pt}
\multirow{2}{*}{Model} & \multicolumn{3}{c}{MAPE in \%}                                  \\ \cline{2-4} 
                       & Bytom               & Decentraland        & Golem               \\ \hline
AGCRN                  & 34.46±1.37          & 26.75±1.51          & 22.83±1.91          \\
Z-GCNETs               & 31.04±0.78          & 23.81±2.43          & 22.32±1.42          \\
STG-NCDE               & 29.65±0.63          & 24.13±1.07          & 22.24±1.53                  \\
TAMP-S2GCNets          & 29.26±1.06          & 19.89±1.49          & \textbf{20.10±2.30}          \\ \hline
\textbf{Localised AGCRN}            & \textbf{28.17±0.93} & \textbf{13.33±0.31} & 21.71±0.49 \\ 
\Xhline{1pt}
\end{tabular}
\label{tab:ether}
\end{table}

\section{Ablation Studies}
\label{sec:discuss}

\subsection{Impact on Resource Efficiency}
As mentioned in \secref{sec:intro}, one of the reasons that we are particularly interested in the localisation of ASTGNNs is its resource efficiency.
Non-localised ASTGNNs usually learn spatial graphs that are complete.
Therefore, the number of edges and consequently the computation overhead grows quadratically with the number of vertices.
Localisation of ASTGNNs is equivalent to pruning edges in the learned spatial graph, which could dramatically reduce the computation overhead associated with the spatial dependencies and thus improve resource efficiency.
To this end, we calculated the amount of computation during inference of 99\%-localised AGCRNs and AGFormers, measured in FLOPs, and summarised the results in \tabref{tab:flops}.
We can see that the localisation of AGCRNs and AGFormers effectively reduces the amount of computation required for inference.
The acceleration is more prominent on AGCRNs against AGFormers because a larger portion of the total computation required by AGFormers is used on their temporal module (transformer layers), whereas AGCRNs use much lighter temporal modules (RNN layers).

\subsection{Localised AGCRNs vs. Other Non-Localised ASTGNNs}
In \figref{fig:pemsd}, \figref{fig:covid} and \figref{fig:ether}, we can clearly see that the localisation up to 99\% is able to improve the test accuracy slightly.
For example, 99\%-localised AGCRNs outperform non-localised AGCRNs by decreasing the RMSE/MAE/MAPE by 3.6\%/3.7\%/2.0\% on PeMSD3.
Such improvement is consistently observed across both AGCRNs and AGFormers and all tested datasets.
We reckon that this improvement is caused by the regularisation effect of sparsifying the spatial graph, which may suggest that the non-localised AGCRNs and AGFormers all suffer from overfitting to a certain degree.

Recent works on ASTGNNs proposed improved architectures of AGCRN, including Z-GCNETs \cite{bib:ICML21:Chen}, STG-NCDE \cite{bib:AAAI22:Choi}, and TAMP-S2GCNets \cite{bib:ICLR22:Chen} for different applications. 
We are therefore curious about how our localised AGCRNs compare to these variations.
Hence we compare the test accuracy of 99\%-localised AGCRNs with these architectures.
Results are shown in \tabref{tab:traffic}, \tabref{tab:covid} and \tabref{tab:ether}.
We can see that our localised AGCRNs can generally provide competitive inference performance even against those delivered by state-of-the-art architectures.
This observation also agrees with our first hypothesis mentioned in \secref{sec:main_results}: in many spatial-temporal datasets, the information provided by the spatial dependencies are primarily included in the information provided by the temporal dependencies.
Therefore, different spatial modules, given that the temporal modules are properly trained, may not make a significant difference in inference performance.

\subsection{Localisation of Non-temporal Graphs}
\label{sec:extend}

To further investigate spatial dependencies and indirectly test our hypotheses, we conduct additional experiments and extend \sysname to non-temporal graphs. 
We attempt to sparsify the spatial graph pre-given by non-temporal datasets, including Cora, CiteSeer, and Pubmed \cite{bib:others08:Sen}. 
On these datasets, we train two non-temporal graph neural network architectures, GCN\cite{bib:ICLR17:Kipf} and GAT\cite{bib:ICLR18:Velickovic}.
On GCN and GAT, as they don't own node embeddings $\mathbf{E}$, we can use the representation $\mathbf{H}$ learned by pre-trained GCN and GAT to replace $\mathbf{E}$ in \eqref{eq:V}, then prune the edges as in \algoref{alg:AGS}, where the weighting factor $\lambda$ takes control of graph sparsity.

\begin{table}[t]
\caption{Classification accuracy (\%) of localised GCN and GAT on citation graph datasets.}
\begin{tabular}{l|llllll}
\Xhline{1pt}
\multirow{2}{*}{Sparsity(\%)} & \multicolumn{2}{l}{Cora} & \multicolumn{2}{l}{Citeseer} & \multicolumn{2}{l}{PubMed} \\ \cline{2-7} 
      & GCN   & GAT   & GCN   & GAT   & GCN   & GAT   \\ \hline
0\%   & 80.20 & 82.10 & 69.40 & 72.52 & 78.90 & 79.00 \\
30\%  & 80.35 & 83.17  & 69.23 & 72.31 & 79.14 & 79.23 \\
50\%  & 72.73 & 75.40 & 69.37 & 72.70 & 78.82 & 79.31 \\
80\%  & 65.19 & 70.81 & 58.47 & 63.18 & 68.37 & 77.03 \\
100\% & 56.22 & 63.29 & 53.13 & 57.50  & 61.02 & 64.25 \\ 
\Xhline{1pt}
\end{tabular}
\label{tab:citation}
\end{table}

\tabref{tab:citation} shows the accuracy of localised GCN and GAT non-temporal datasets. The pre-defined graphs of Cora, Citeseer, and PubMed are sparsified to 30\%, 50\%, 80\% and 100\%, respectively.
We can observe a significant drop in the test accuracy among all localised non-temporal graph models.
This indicates that, in the absence of temporal dependencies, the information provided by spatial dependencies plays a major role during inference and thus can not be ignored via localisation.


\section{Conclusion}
\label{sec:conclusion}
In this paper, we ask the following question: whether and to what extent can we localise spatial-temporal graph models?
To facilitate our investigation, we propose \sysname, a novel algorithm dedicated to the sparsification of adjacency matrices in ASTGNNs.
We use \sysname to localise two ASTGNN architectures: AGCRN and AGFormer, and conduct extensive experiments on nine different spatial-temporal datasets.
Primary experiment results showed that The spatial adjacency matrices could be sparsified to over 99.5\% without deterioration in test accuracy on all datasets.
Furthermore, when the ASTGNNs are fully localised, we still observe no accuracy drop on the majority of the tested datasets, while only minor accuracy deterioration happened on the rest datasets.
Based on these observations, we suggest two hypotheses regarding spatial and temporal dependencies: \textit{(i)} in the tested data, the information provided by the spatial dependencies is primarily included in the information provided by the temporal dependencies and, thus, can be essentially ignored for inference;
and \textit{(ii)} although the spatial dependencies provide redundant information, it is vital for effective training of ASTGNNs and thus cannot be ignored during training.
Last but not least, we conduct additional ablation studies to show the practical impact of ASTGNN's localisation on resource efficiency and to verify our hypotheses from different angles further.

\section{Acknowledgement}
The authors are grateful to the KDD anonymous reviewers for many insightful suggestions and engaging discussion which improved the quality of the manuscript.

\clearpage
\bibliographystyle{ACM-Reference-Format}
\bibliography{cites}

\appendix
\section{Appendix}

\begin{table*}[t]
\caption{Dataset-specific hyperparameter setup for AGCRN and AGFormer.}
\footnotesize
\begin{tabular}{ll|cc|cc|cc|cc}
\Xhline{1pt}
\multirow{2}{*}{Task type} &
  \multirow{2}{*}{Datasets} &
  \multicolumn{2}{l|}{hidden dimension} &
  \multicolumn{2}{l|}{embedding dimension} &
  \multicolumn{2}{l|}{batch size} &
  \multicolumn{2}{l}{learning rate} 
  \\ 
  \cline{3-10}
         &              & AGCRN & AGFormer & AGCRN & AGFormer & AGCRN & AGFormer & AGCRN & AGFormer \\ 
 \hline
Traffic  & PeMSD3       & 64    & 32       & 5   & 2       & 64    & 128      & 3e-3  & 5e-4    \\
Traffic  & PeMSD4       & 64    & 32       & 2    & 5       & 64    & 128      & 3e-3  & 5e-4   \\
Traffic  & PeMSD7       & 64    & 32       & 2    & 2        & 64    & 128      & 3e-3  & 5e-4    \\
Traffic  & PeMSD8       & 64    & 32       & 5    & 5      & 64    & 128      & 3e-3  & 5e-4    \\
Ethereum & Decentraland & 32    & 32       & 1     & 2        & 8     & 32       & 1e-2  & 1e-3    \\
Ethereum & Bytom        & 32    & 32       & 1     & 2        & 8     & 32       & 1e-2  & 1e-3   \\
Ethereum & Golem        & 32    & 32       & 1     & 2        & 8     & 32       & 1e-2  & 1e-3     \\
COVID-19 & CA           & 128   & 64     & 5    & 2       & 16    & 64       & 6e-3  & 2e-3     \\
COVID-19 & TX           & 128   & 64      & 5    & 5       & 16    & 64       & 6e-3  & 2e-3    \\ 
\Xhline{1pt}
\end{tabular}
\label{tab:configure}
\end{table*}

\subsection{Additional Dataset Details}\label{app:datasets}
The detailed datasets and configurations are as follows:

\begin{itemize}
    \item \textbf{Transporation}: We use four widely-studied traffic forecasting datasets from Caltrans Performance Measure System (PeMS): \textbf{PeMSD3}, \textbf{PeMSD4}, \textbf{PeMSD7} and \textbf{PeMSD8} \cite{bib:others01:Chen}. Following \cite{bib:NIPS20:Bai}, PeMSD3, PeMSD4, PeMSD7, and PeMSD8 are split with a ratio of 6:2:2 for training, validation and
    testing. The traffic flows are aggregated into 5-minute intervals. We conduct 12-sequence-to-12-sequence forecasting, the standard setting in this domain. The accuracy is measured in Mean Absolute Error (MAE), Root Mean Square Error (RMSE) and Mean Absolute Percentage Error (MAPE).
    \item \textbf{Blockchain}: We use three Ethereum price prediction data-sets: \textbf{Bytom}, \textbf{Decentral} and \textbf{Golem}\cite{bib:SMD20:Li}. 
    These data are represented in graphs, with nodes and edges being the addresses of users and digital transactions, respectively. 
    The interval between two consecutive time points is one day. 
    Following \cite{bib:ICLR22:Chen}, Bytom, Decentraland, and Golem token networks are split with a ratio of 8:2 for training and testing. 
    We use 7 days of historical data to predict a future of 7 days. MAE, RMSE, and MAPE are also used as accuracy metrics.
    \item \textbf{Biosurveillance}: We use California (\textbf{CA}) and Texas (\textbf{TX}), COVID-19 biosurveillance datasets\cite{bib:ICLR22:Chen} used to forecast the number of hospitalized patients. 
    The time interval of these datasets is one day.
    Following \cite{bib:ICLR22:Chen}, CA and TX are split with a ratio of 8:2 for training and testing. Three days of historical data are used to predict a future of fifteen days.  
    With these two datasets, MAE and RMSE are not used for measuring the inference accuracy, as their values are extremely small and, thus, difficult to reflect the performances realistically. 
    Following \cite{bib:ICLR22:Chen}, we only use MAPE.
\end{itemize}

\subsection{Experiment Details}\label{app:detail}
The dataset-specific hyperparameters chosen for AGCRNs and AGFormers are summarised in \tabref{tab:configure}. All experiments are implemented in Python with Pytorch 1.8.2 and executed on a server with one NVIDIA RTX3090 GPU. We optimize all the models using the Adam optimizer\cite{bib:ICLR15:Ma}. 
For transportation data, we set the number of pre-training iterations $\mathrm{N}_1=150$, and the number of sparsification iterations $\mathrm{N}_2=400$. We use an early stop strategy with a patience of 15 for pre-training. Parameters are chosen through a parameter-tuning process on the validation set. For blockchain data, we set the number of pre-training iterations $\mathrm{N}_1=100$, and the number of sparsifcation iterations $\mathrm{N}_2=200$. We use an early stop strategy with a patience of 30 for pre-training. Parameters are chosen through a parameter-tuning process on the training set. For biosurveillance data,  we set the number of pre-training iterations $\mathrm{N}_1=100$, and the number of sparsifcation iterations $\mathrm{N}_2= 200$. We use an early stop strategy with a patience of 15 for pre-training. Parameters are chosen through a parameter-tuning process on the training set.
\subsection{Notation}
Frequently used notations are summarized in \tabref{tab:notation}.

\begin{table}[htbp]
\centering
\caption{The main symbols and definitions in this paper.}
\begin{tabular}{ll}
\Xhline{1pt}
Notation&Definition \\ \hline
$\mathcal{G}$& the spatial network  \\
$N$&the number of nodes  \\
$\mathcal{X}$&a sequence of discrete frames\\
$\mathcal{T}$&length of historical observations\\
$\mathcal{H}$&length of future observations\\
$\boldsymbol{X}_{t}$&  node feature matrix at timestep $t$\\
$\mathbf{E}$ & the learnable node embedding \\
$d$ &the node embedding dimension  \\
$C$&the node feature dimension  \\
$F$& the feature dimension\\
$\mathbf{A}_{a d p}$&the normalized adaptive adjacency matrix\\
$\mathbf{M}_{\mathbf{A}}$&mask to prune $\mathbf{A}_{a d p}$  \\
$\mathcal{B}$ &the Bernoulli distribution  \\
 $\mathcal{F}(\cdot)$&the ASTGNN model  \\
$\mathrm{N}_1$&number of  pre-training iterations  \\
 $\mathrm{N}_2$& number of sparsifcation iterations \\\Xhline{1pt}
\end{tabular}
\label{tab:notation}
\end{table}

\end{document}